\documentclass[10pt, a4paper]{article}
\usepackage{lrec2022} 
\usepackage{multibib}
\usepackage{todonotes}
\usepackage{graphicx}
\usepackage{tabularx}
\usepackage{soul}
\usepackage{titlesec}
\titleformat{\section}{\normalfont\large\bfseries\center}{\thesection.}{1em}{}
\titleformat{\subsection}{\normalfont\SmallTitleFont\bfseries\raggedright}{\thesubsection.}{1em}{}
\titleformat{\subsubsection}{\normalfont\normalsize\bfseries\raggedright}{\thesubsubsection.}{1em}{}
\renewcommand\thesection{\arabic{section}}
\renewcommand\thesubsection{\thesection.\arabic{subsection}}
\renewcommand\thesubsubsection{\thesubsection.\arabic{subsubsection}}

\usepackage{epstopdf}
\usepackage[utf8]{inputenc}

\usepackage{hyperref}
\usepackage{xstring}

\usepackage{color}

\usepackage{caption}
\usepackage{subcaption}

\usepackage{fancyvrb} 

\title{How much of UCCA can be predicted from AMR?}

\name{Siyana Pavlova, Maxime Amblard, Bruno Guillaume} 

\address{Université de Lorraine, CNRS, Inria, LORIA, F-54000 Nancy, France \\
         \{firstname.lastname\}@loria.fr\\}

\abstract{
In this paper, we consider two of the currently popular semantic frameworks: Abstract Meaning Representation (AMR) - a more abstract framework, and Universal Conceptual Cognitive Annotation (UCCA) - an anchored framework. We use a corpus-based approach to build two graph rewriting systems, a deterministic and a non-deterministic one, from the former to the latter framework. We present their evaluation and a number of ambiguities that we discovered while building our rules. Finally, we provide a discussion and some future work directions in relation to comparing semantic frameworks of different flavors. 
 \\ \newline \Keywords{Semantic Framework, Graph Rewriting, Abstract Meaning Representation, Universal Conceptual Cognitive Annotation} }

\begin{document}

\maketitleabstract

\section{Introduction and Motivation\label{sec:intro}}

A number of frameworks for semantic annotation have been proposed in the past decades. As each puts the main focus on a different aspect of semantics, each is fit for its purpose, has its set of adopters and there is no one framework that is better than the rest. As a result, semantically annotated data, which is not easy to come by in the first place and is laborious and time-consuming to produce manually, is scattered across different frameworks. It would be useful if we can transform annotations from one framework into another, thus making more data available in various frameworks.

In the current work, we focus on the comparison between two of the existing semantic frameworks, with different relations to anchoring - one anchored and one more abstract - and an experiment we carried out to see how much of the former can be predicted from the latter. These frameworks are Universal Conceptual Cognitive Annotation (UCCA)~\cite{abend2013universal} and Abstract Meaning Representation (AMR)~\cite{banarescu2013abstract}.

In \autoref{sec:background}, we give an overview of the two frameworks that we consider in this work as well as the shared task from which the data we use comes from. In \autoref{sec:experiment}, we describe the Graph Rewriting experiment we carried out to transform AMR graphs into UCCA-like structures. Then \autoref{sec:evaluation} describes how our graph rewriting system was evaluated and reports our results and observations. In \autoref{sec:case_study} we present some of the ambiguous cases we discovered when building out rewriting systems. Finally, in \autoref{sec:conclusion}, we provide a broader discussion on some of the points stemming from this experiment and some future work directions.

\section{Background\label{sec:background}}

Our choice of frameworks is grounded in the current popularity of the two we are considering - AMR is often discussed in the community, with proposals for potential enhancements in many of the semantic workshops and conferences, and UCCA has increasingly been gaining traction in the past years, with more data being made available continuously and proposals for extension layers being made too.

Additionally, AMR and UCCA are two of the frameworks that were part of the 2019 and 2020 Meaning Representation Parsing (MRP) shared tasks~\cite{oepen-etal-2019-mrp,oepen-etal-2020-mrp} thanks to which there is parallel annotated data for the two, even though only a small amount (87 sentences from the WSJ corpus) is freely available.

\subsection{AMR\label{subsec:amr_background}}

\begin{figure}
  \centering
\begin{small}
\begin{verbatim}
(a / adjust-01
   :ARG0 (b / girl)
   :ARG1 (m / machine))
\end{verbatim}
\end{small}

  \includegraphics[scale=.35]{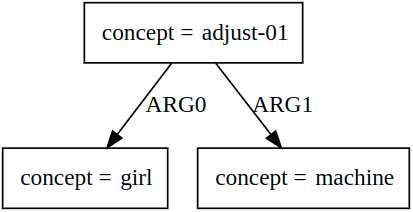}
  \caption{AMR annotation of the sentences ``\emph{The girl made adjustments to the machine.}'', 
``\emph{The girl adjusted the machine.}'' and
``\emph{The machine was adjusted by the girl.}'' in Penman format (top) and as a graph (bottom).}
  \label{amr_spec_adjust}
\end{figure}

AMR was introduced in 2013. Broadly speaking, it represents ``who did what to whom'' in a sentence. 
AMR abstracts from the surface representation of a sentence and is what~\cite{koller-etal-2019-graph} describe as a flavor 2 semantic framework, where the ``flavor'' of a framework stands for correspondence between surface level tokens and graph nodes. In flavor 2 frameworks, such as AMR, there is no direct correspondence between the two - not all tokens are present as nodes in the graph and not all graph nodes correspond to tokens. Thus, sentences that are different on the surface, but have the same basic meaning are represented by the same AMR. For example, the AMR in \autoref{amr_spec_adjust} is the representation of the sentence
``\emph{The girl made adjustments to the machine.}'', but also of the sentences
``\emph{The girl adjusted the machine.}'' and ``\emph{The machine was adjusted by the girl.}'' as shown in the official AMR specifications\footnote{\url{https://github.com/amrisi/amr-guidelines/} (at the time of writing, this link points to version 1.2.6 of the specifications)} .

AMR relies heavily on predicate-argument structure and makes extensive use of PropBank predicates~\cite{palmer-etal-2005-proposition}, trying to maximize their use whenever possible in sentences. Predicates are used not only to annotate the verbs in a sentence, but also the nouns and adjectives whenever possible. As seen with the example from \autoref{amr_spec_adjust}, the noun \emph{adjustment} and the verb \emph{adjust} are both annotated with the PropBank predicate \texttt{adjust-01}. The arguments of PropBank predicates appear as core roles in AMR graphs. In addition, non-core roles such as \texttt{location}, \texttt{time}, \texttt{purpose}, etc. form the rest of the AMR relations.

In terms of graph features, AMR graphs are directed acyclic graphs (DAGs) and singly-rooted. The acyclicity and single-rootedness come at the cost of using inverse relations. Any role, core or non-core, can be reversed by adding \texttt{-of} to its name and changing the direction of the relation. Apart from avoiding cycles, inverse roles also serve to highlight the \textit{focus} of a sentence by making sure that the central concept is the root of the AMR graph.

The AMR Bank is a manually-produced corpus of AMR annotations in English. Only a portion of it (namely the Little Prince corpus and the BioAMR corpus) are freely available. The rest of the AMR Bank can be obtained by a (paid) license from the Linguistics Data Consortium.
AMR was designed with English in mind and does not aim to be a universal semantic representation framework. That being said, there have been attempts to use the framework for other languages, notably Chinese, in the Chinese AMR (CAMR) Bank\footnote{\url{https://www.cs.brandeis.edu/~clp/camr/camr.html}}.

While powerful in its ability to abstract from surface representation, there are a number of phenomena that the framework does not cover - tense, plurality, definiteness, scope, to name some of the more prominent ones. Some of these issues have been addressed: \cite{bos-2020-separating} proposes an extension to deal with scope in AMR, while~\cite{donatelli-etal-2018-annotation} proposes to augment AMR with tense and aspect. However, to the best of our knowledge, no corpora exist that use the proposed extensions yet.

\subsection{UCCA\label{subsec:ucca_background}}

UCCA was introduced in 2013 as well, but has gained more traction in recent years - a number of extension layers have been proposed and the number of available annotated datasets has been increasing.

Following~\cite{koller-etal-2019-graph}'s flavor classification, UCCA is a flavor 1 framework, i.e. an \emph{anchored} framework - each token (or a group of tokens in the case of named entities, such as proper names and dates) corresponds to a leaf node in the graph, but additional nodes are present in the graph too. UCCA organises processes (actions) and states into \textit{scenes}, where the central process or state, its participants, temporal and adverbial information are labeled. Each of these may expand into its own subgraph where elaborations, quantifiers, function and relation words are labeled. A sentence may give rise to multiple scenes and these can be labeled as well. UCCA offers 14 relation types in total. It allows for re-entrances via the so-called 
``remote'' edges. As with AMR, UCCA graphs are also singly-rooted DAGs.

\begin{figure}
\centering

  \includegraphics[scale=.27]{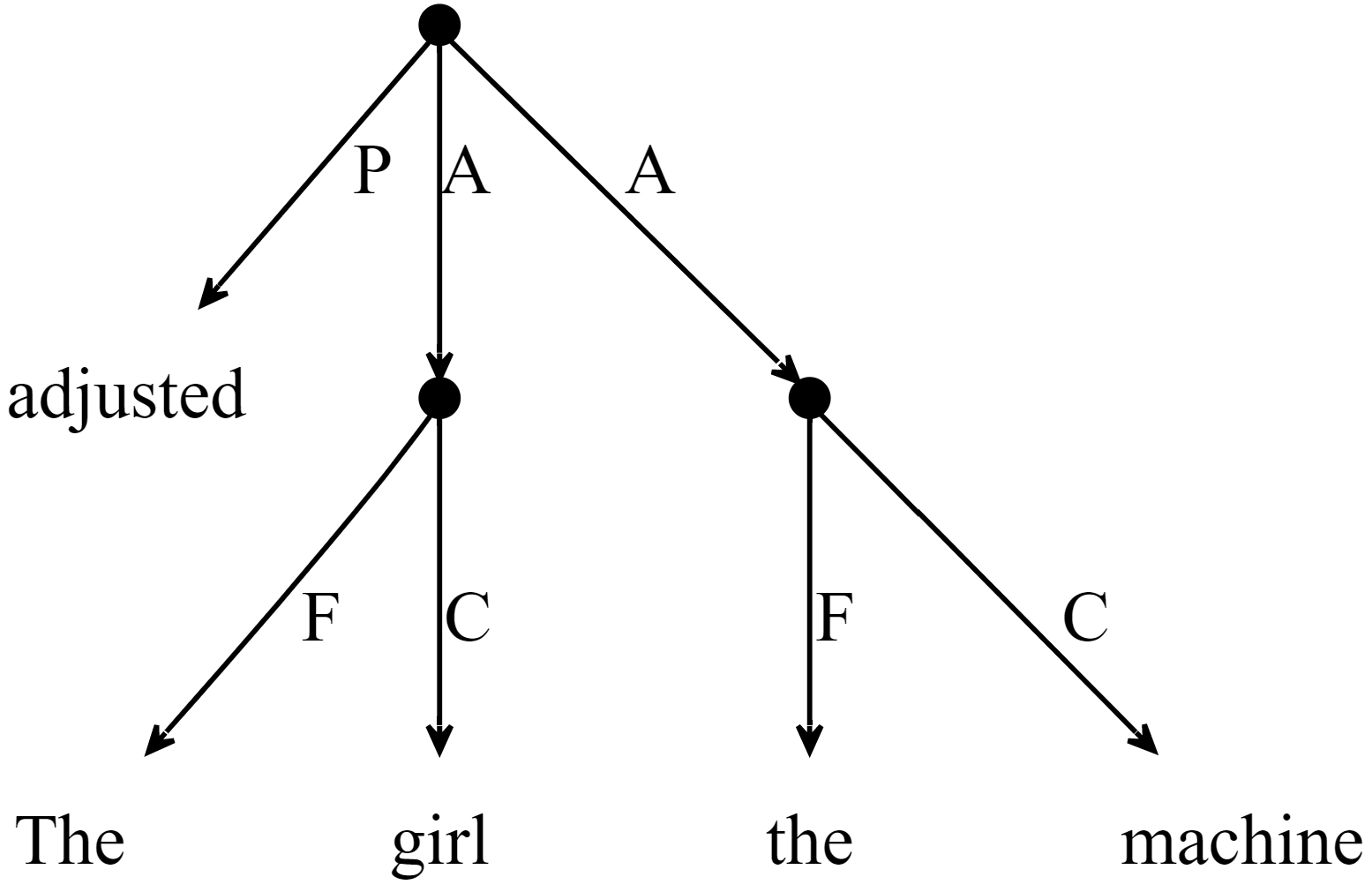}
  
\begin{small}
\begin{Verbatim}[commandchars=\\\{\},codes={\catcode`$=3\catcode`_=8}]

[The\textbf{\textsubscript{F}} girl\textbf{\textsubscript{C}}]\textbf{\textsubscript{A}} adjusted\textbf{\textsubscript{P}} [the\textbf{\textsubscript{F}} machine\textbf{\textsubscript{C}}]\textbf{\textsubscript{A}}
\end{Verbatim}
\end{small}
  
  \caption{UCCA annotation of the sentence ``\emph{The girl adjusted the machine.}'' as a graph (top) and in textual format (bottom).}
  \label{ucca_adjust}
\end{figure}

\autoref{ucca_adjust} shows the UCCA annotation of the sentence ``\emph{The girl adjusted the machine}''. The process \texttt{P} at the center of the scene is \textit{adjusted}. That scene includes two participants \texttt{A}, which are internally annotated further, with the central concept (\textit{girl}, \textit{machine}) receiving the label \texttt{C} and the function word \textit{the} - \texttt{F}.

As the name suggests, UCCA is designed to be a \emph{universal} semantics framework, i.e. it aims to work across languages as opposed to being designed with a specific language in mind. Indeed, the currently available UCCA datasets span across English, French, German and Hebrew.

A number of extension layers have been proposed for UCCA, such as adding semantic roles~\cite{shalev-etal-2019-preparing,prange-etal-2019-made}, co-reference~\cite{prange-etal-2019-semantically} and implicit arguments~\cite{cui-hershcovich-2020-refining}. However, while small proof of concept datasets exist for some of these, there is no parallel corpus between any of the UCCA extension layers and other semantic frameworks, such as AMR. Therefore, for this study we concentrate on the foundational layer of UCCA. 

\subsection{MRP\label{subsec:mrp_background}}

The MRP 2019 and 2020 Shared Tasks are parsing tasks, that have sentences annotated in a number of semantic frameworks. AMR, UCCA, DM: DELPH-IN MRS Bi-Lexical Dependencies (DM), Prague Semantic Dependencies (PSD) and Elementary Dependency Structures (EDS) were part of the 2019 task. The 2020 task drops DM and PSD in favour of Prague Tectogrammatical Graphs (PTG) and Discourse Representation Graphs (DRG). All the sentences in these datasets are in English. Both tasks use the same portion of the WSJ corpus in the freely available sample\footnote{\url{http://svn.nlpl.eu/mrp/2019/public/sample.tgz}} of annotations and so for the purposes of comparing AMR and UCCA, they are equivalent. The sample contains an overlap of 87 annotated sentences for both AMR and UCCA, which we have used for this study.

An evaluation tool, \textit{mtool}\footnote{\url{https://github.com/cfmrp/mtool}}, was introduced for these tasks as well and is what we make use of for our evaluation.

It must be noted that the UCCA graphs are not entirely consistent with the UCCA guidelines\footnote{\url{https://github.com/UniversalConceptualCognitiveAnnotation/docs/releases)} (at the time of writing this link points to v2.1 of the guidelines}. There are a few small structural differences, which can easily be adjusted, but our analysis, especially when discussing the \textit{mtool} evaluation scores, will be misleading without highlighting these differences. These are (1) punctuation is not annotated in the guidelines, but is in the MRP dataset and (2) the root node from the UCCA guidelines would not be the same as the one in the MRP dataset.

The MRP graphs for AMR are generally consistent with the AMR specifications. With that being said, we have discovered on error in the annotations. The AMR specifications state that ``to represent conjunction, AMR uses concepts \texttt{and}, \texttt{or}, \texttt{contrast-01}, \texttt{either}, and \texttt{neither}, along with \texttt{:opx} relations''. We note that sentence \texttt{\#20003008} has not been annotated in the best possible way because the annotation uses \texttt{and} plus \texttt{:polarity -} (see \autoref{lorillard_amr}) when \texttt{neither} is available and arguably a more appropriate option.

\section{Experiments\label{sec:experiment}}

\subsection{Data and Data Processing}

As mentioned in \autoref{subsec:mrp_background}, we use the freely available sample of annotations from the MRP 2019 and 2020 Shared Tasks. The corpus has 87 sentences that overlap between UCCA and AMR. We use the first 17 sentences (called the train set hereupon), which constitute 20\% of the corpus, to construct the rules for our graph rewriting system. The remaining 70 sentences are our test set, used for evaluation.

\begin{figure*}[t!]
  \centering
  
  \begin{subfigure}[b]{0.39\linewidth}
    \centering
    \includegraphics[scale=.26]{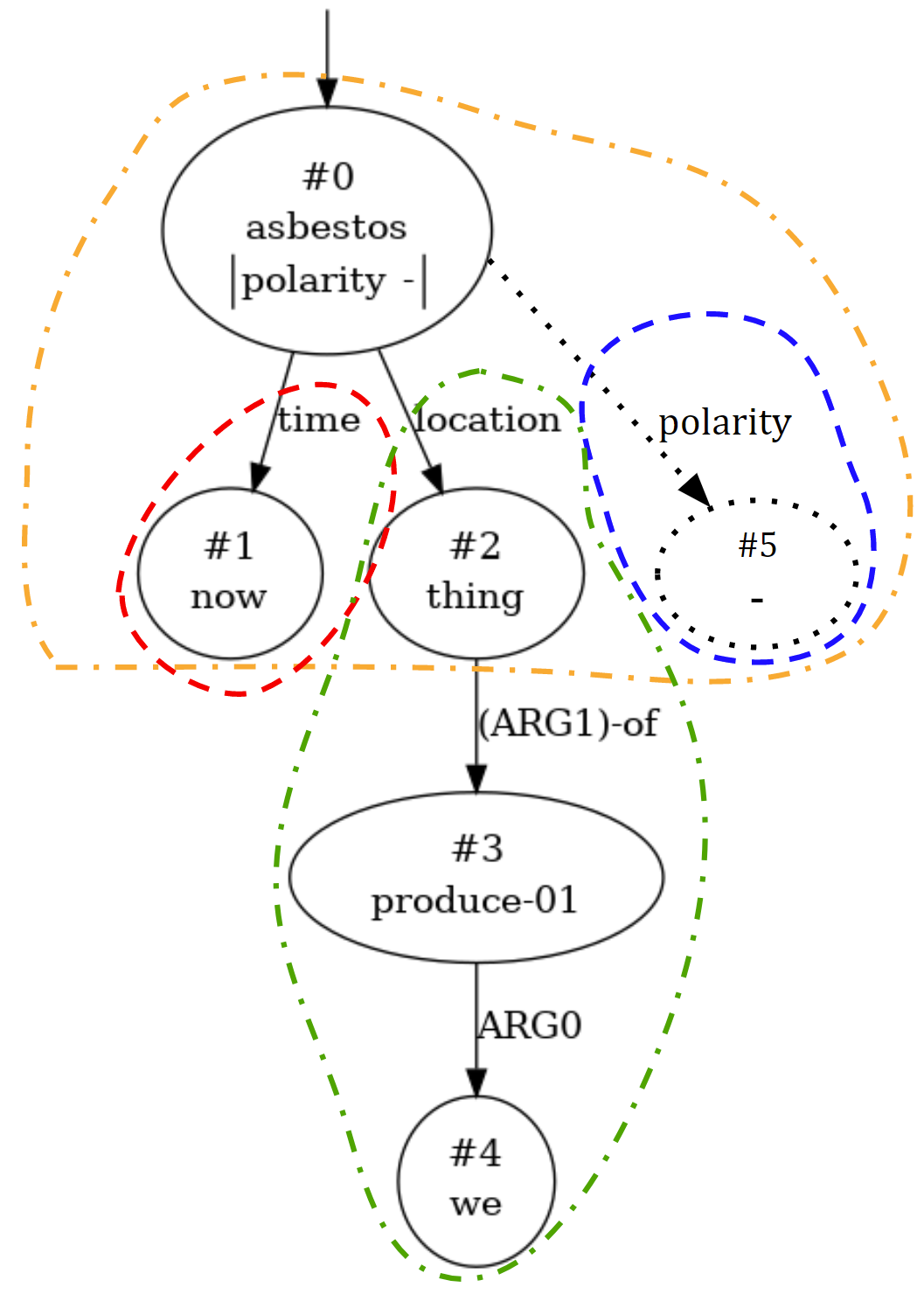}
    \caption{AMR with property-value pair \texttt{polarity -} extracted as an edge and a node.}
    \label{asbestos_amr}
  \end{subfigure}
  \begin{subfigure}[b]{0.60\linewidth}
    \centering
    \includegraphics[scale=.27]{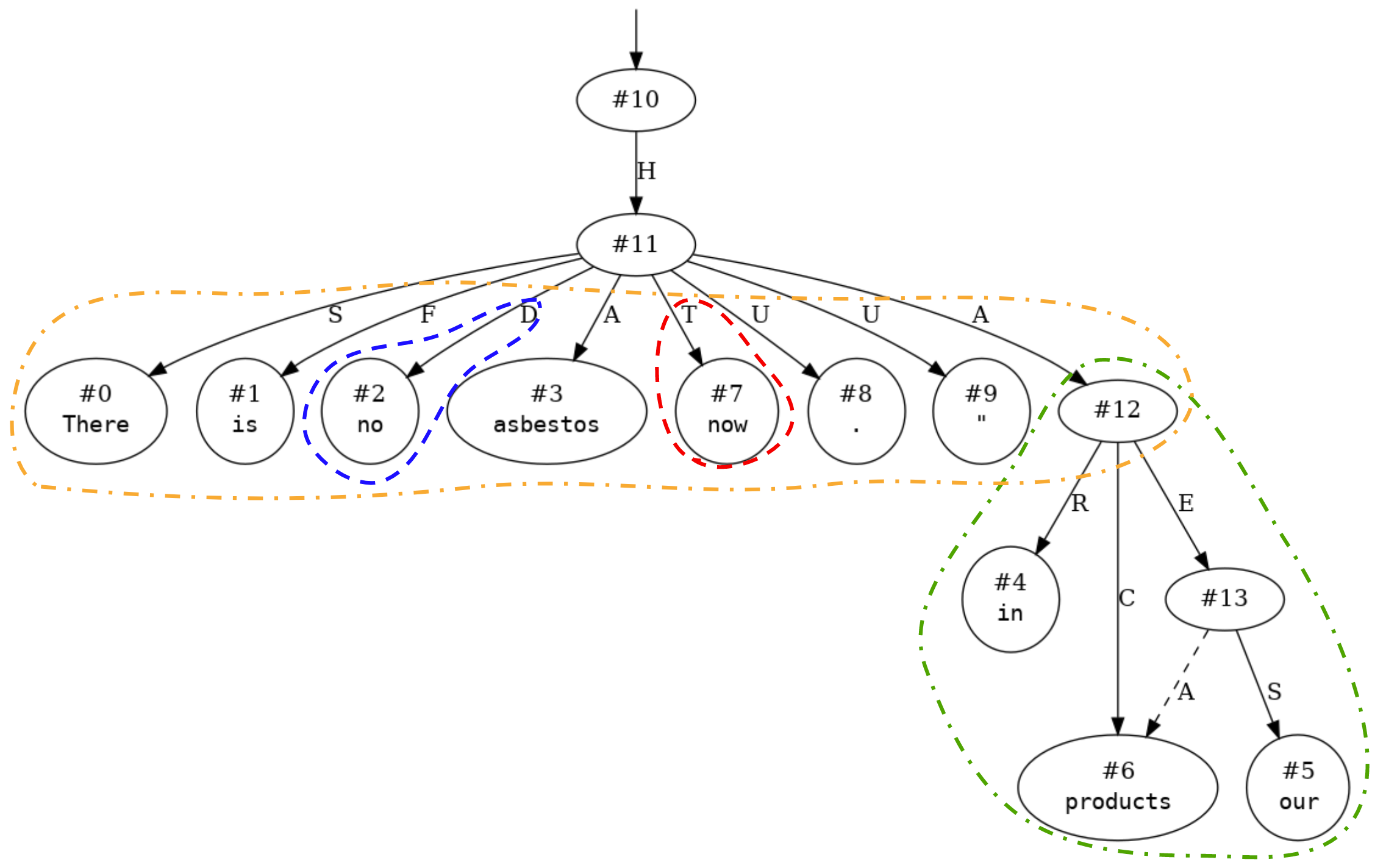}
    \caption{UCCA}
    \label{asbestos_ucca}
  \end{subfigure}
  
  \caption{AMR and UCCA annotations of the sentence ``\emph{There is no asbestos in our products now.''}'' \texttt{[\#20003007]} with corresponding subgraphs circled in matching colours.}
  \label{asbestos}
\end{figure*}

The data in the shared task is provided both in \texttt{JSON} and in \texttt{DOT} format. \texttt{PDF} files with the graphs generated from the DOT files are also provided. We used the aforementioned \texttt{DOT} files to produce images of the two graphs (AMR and UCCA) for each sentence alongside each other. The AMR graphs were then manually adjusted so that property-value pairs were turned into edges and nodes, as in many cases the values directly corresponded to UCCA nodes and made it more straightforward to draw parallels between the two representations. For example, for sentence \texttt{\#20003007} (\autoref{asbestos_amr}), the property-value pair \texttt{polarity -} of node \texttt{\#0}, was transformed to an edge \texttt{polarity} from node \texttt{\#0} to a new node with label \texttt{-} and given the next available ID number (\texttt{\#5}). Comparing that with the UCCA graph of the same sentence in \autoref{asbestos_ucca}, we can see these new node and edge directly correspond to node \texttt{\#2} labeled \texttt{no} and its incoming \texttt{D} edge.

We used these modified pictorial representations of the graphs to make our first observations. For each sentence, we manually identified the corresponding (overlapping) subgraphs between the AMR graph and the UCCA graph. As a rule, we marked subgraphs as sets of predicates along with their arguments and any properties of the arguments (e.g. \texttt{opN}, \texttt{year}, \texttt{month}). Furthermore, clearly identifiable direct transformations between relations were marked. For example, in the example in \autoref{asbestos}, \texttt{time} and \texttt{polarity} can be directly linked to \texttt{T} and \texttt{D} respectively\footnote{The coloured pictures for the 17 sentences along with the code and data for the experiments are available at \url{https://gitlab.inria.fr/semagramme-public-projects/resources/amr2ucca}}. Through this we made some initial observations about the most probable correspondents for each AMR relation. We also noted some observations about the differences in the generic structure of the graphs. UCCA graphs, unsurprisingly, tend to have more nodes than AMR graphs. In AMR, predicates are parent nodes of their arguments, whereas in UCCA, participants in a scene appear as siblings of the process or state that is at the center of that scene.

\subsection{Graph Rewriting}

We use \textsc{Grew}\footnote{\url{https://grew.fr/}} for graph rewriting~\cite{guillaume-2021-graph,bonfante2018application} from AMR to an UCCA-like structure. \textsc{Grew} allows us to define rules that match patterns in a graph and apply commands to transform the matched part of the graph.

We design two sets of rules. \textbf{R1} is our initial set of rules, which serves as a base line system with a direct and deterministic set of rules. We then build \textbf{R2} - an extended set of rules that tries to cover some of the identified problems with \textbf{R1}, namely (a) more complex structures and (b) ambiguous transformations, for which we use a non-deterministic set of rules.

\subsubsection{Initial set of rules - R1}

We built a set of rules \textbf{R1} based on our initial observations. \textbf{R1} was constructed such that any core and non-core AMR relation was rewritten to its most probable correspondent based on the observation of the train set. Additionally, the AMR root (usually a predicate) was ``pushed down'' to the level of its arguments. Inverse relations were not dealt with separately at this stage. 
\textit{mtool} runs only if all edges in a graph are valid relations from the framework being tested. Therefore, to be able to apply it on the produced graphs, we added a back-off rule, \texttt{ensure\_ucca\_edges}, that rewrites any remaining non-UCCA edges to \texttt{A} (participant). 
We chose \texttt{A} since this was the most frequent relation in the UCCA train set and the relations affected by this rule were mostly \texttt{ARGx-of} relations, where \texttt{x} is the argument number. This also ensures that if there are any relations in the test set that were not present in our train set, they will still be transformed into a valid UCCA relation.

\begin{figure*}
  \centering
  \begin{subfigure}[b]{0.35\linewidth}
\begin{small}
\begin{verbatim}
rule time_to_T {
    pattern {
        e: X -[time]-> Y;
    }
    commands {
        del_edge e;
        add_edge X -[T]-> Y;
    }
}
\end{verbatim}
\end{small}
\caption{}
\label{r1_time_to_T_rule}
\end{subfigure}
    \begin{subfigure}[b]{0.30\linewidth}
        \centering
        \includegraphics[scale=.23]{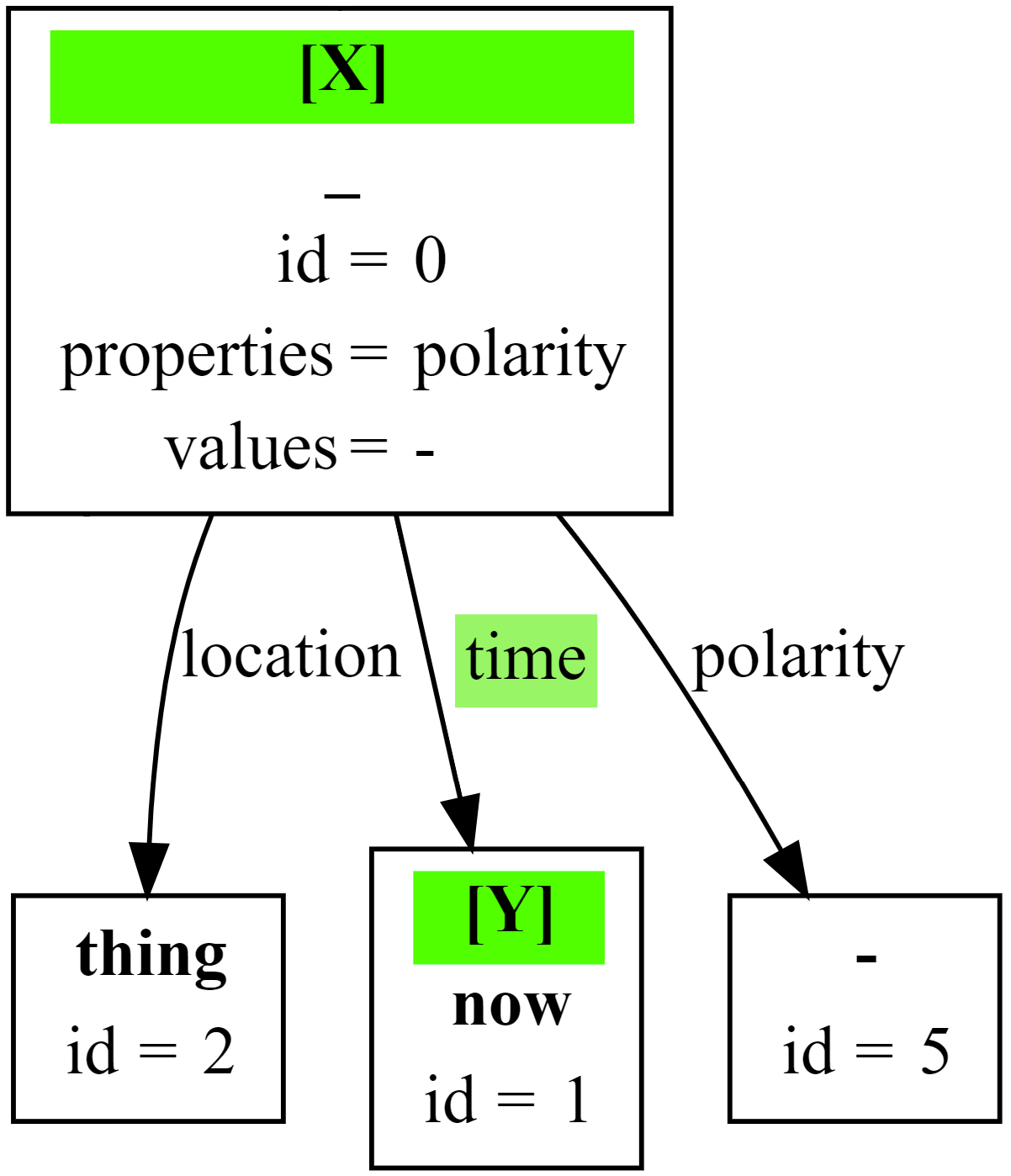}
        \caption{}
        \label{r1_time_to_T_matching}
    \end{subfigure}
    \begin{subfigure}[b]{0.30\linewidth}
        \centering
        \includegraphics[scale=.23]{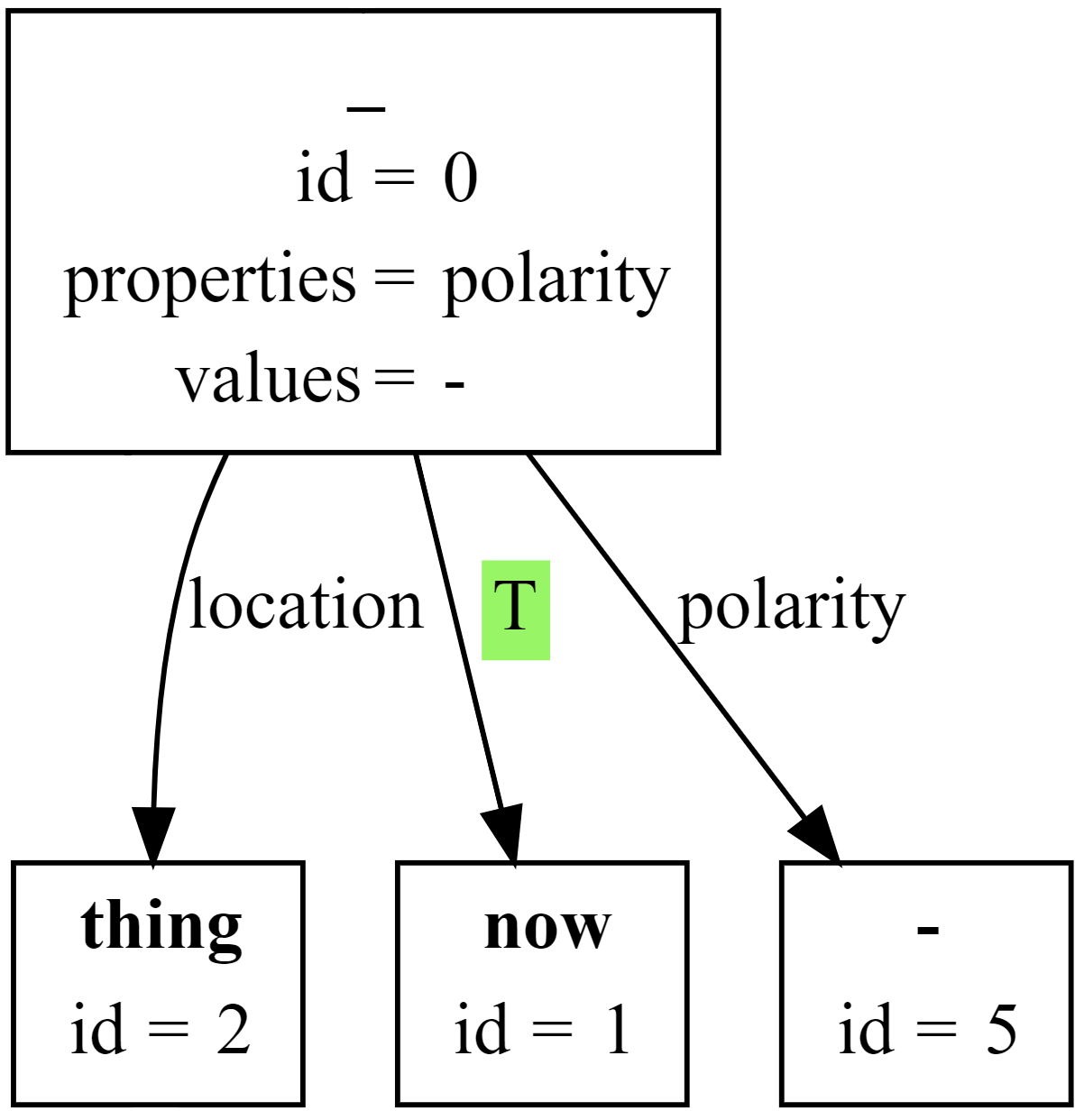}
        \caption{}
        \label{r1_time_to_T_rewriting}
    \end{subfigure}

  \label{r1_time_to_T}
  \caption{The rule \texttt{time\_to\_T} (a), the subgraph of the sentence ``\emph{There is no asbestos in our products now.''}'' \texttt{[\#20003007]} that it matches (b) and the resulting subgraph after rewriting (c).}
\end{figure*}

\autoref{r1_time_to_T_rule} shows one of the rules in R1, \texttt{time\_to\_T} which matches a pair of nodes that are linked via the AMR relation \texttt{time} and the edge itself. If such a pattern is found, the rule deletes the \texttt{time} relation and adds a \texttt{T} relation from the parent to the child. In \autoref{r1_time_to_T_matching}, highlighted in green, we can see the part of the graph for sentence \texttt{\#20003008} of the corpus that has been matched by this rule. In \autoref{r1_time_to_T_rewriting}, we see the resulting subgraph after the rewriting.

\texttt{time\_to\_T} is one of the 16 rules that constitute \textbf{R1}. The first rule, \texttt{push\_root\_down}, is applied once at the start. It puts the sentence in a parallel scene (\texttt{H}) in order to comply with the dataset structure. Other rules are then iterated as much as possible. Finally the back-off rule rewrites any remaining non-UCCA edges to \texttt{A}. 

\subsubsection{Extended set of rules - R2}

Next, we constructed \textbf{R2} - an extension of \textbf{R1}, following a more systematic approach. Each of the AMR relations, along with special AMR nodes (e.g. \texttt{have-org-role-91}) present in the corpus\footnote{There are 27 relations in the first 17 sentences of the corpus: \texttt{ARG0}, \texttt{ARG1}, \texttt{ARG2}, \texttt{ARG3}, \texttt{ARG4}, \texttt{ARG4}, \texttt{day}, \texttt{month}, \texttt{year}, \texttt{domain}, \texttt{mod}, \texttt{name}, \texttt{time}, \texttt{prep-in}, \texttt{location}, \texttt{op1}, \texttt{op2}, \texttt{op3}, \texttt{quant}, \texttt{purpose}, \texttt{decade}, \texttt{part}, \texttt{duration}, \texttt{unit}, \texttt{polarity}, \texttt{topic}, \texttt{manner}, \texttt{age} and \texttt{poss}, \texttt{consist-of}, and seven reversed relations: \texttt{ARG0-of}, \texttt{ARG1-of}, \texttt{ARG2-of}, \texttt{quant-of}, \texttt{polarity-of}, \texttt{part-of}. Though, arguably, \texttt{mod} can be considered as the reverse relation \texttt{domain-of}.} was explored further and either (a) rules were written that account for each of the occurrences of that structure or (b) a conclusion was reached that a specific structure is too ambiguous to rewrite in a decisive manner.

\textbf{R2} contains 44 rules, which, aside from treating the relations from \textbf{R1}, also treat more complex constructions such as conjunction and some special nodes such as \texttt{date-entity}. Furthermore, for two pairs of rules, (\texttt{time\_to\_T}, \texttt{time\_to\_D}) and (\texttt{quant\_to\_D}, \texttt{quant\_to\_Q}), we apply a non-deterministic \textsc{Grew} strategy. This means that whenever faced with a choice between multiple ways to rewrite a relation, the system produces a graph for each possible option and the rest of the rules are applied to each of these, resulting in multiple outputs for a single input graph.

\section{Evaluation\label{sec:evaluation}}

We use \textit{mtool} for the initial evaluation of \textbf{R1} and \textbf{R2}, so that our results are comparable to the systems that participated in the MRP 2019 and 2020 tasks. We report the results in \autoref{tab:r1_r2_results}. We use \textit{mtool}'s \texttt{mrp} setting for \texttt{--score} which, for UCCA graphs, counts the number of anchors, edges, attributes (which in UCCA account for remote edges) and top nodes to compute precision, recall and F1-score.

In \autoref{tab:r1_r2_results}, we present the precision, recall and F1-score for both the train and the test set. Since for \textbf{R2}, we have multiple output graphs per sentence, the scores presented there are the macro-average, i.e. for each sentence, we compute the average value for each metric across all outputs for that sentence, and then average that value across sentences. For the train set, we get $4.05$ output graphs per sentence on average, and for the test set, $2.67$.

While the results are low as such, it is still important to note that they double for our train set and increase significantly for our test set. It is interesting to note that with the exception of precision for \textbf{R2}, our scores are higher on the test set than on the train set. This seems surprising, as one normally expects the opposite to be true. However, with such a small dataset, it is difficult to say whether this is a valid trend or simply due to a non-uniform train-test split.

\begin{table*}[]
    \centering
    \begin{tabular}{c||c|c|c||c|c|c}
         &  \multicolumn{3}{c||}{Train} & \multicolumn{3}{c}{Test}\\
         \hline
         & Precision & Recall & F1-score & Precision & Recall & F1-score\\
         \hline
         \hline
         R1 & 0.128 & 0.037 & 0.057 & 0.173 & 0.055 & 0.083 \\
         R2 & 0.249 & 0.079 & 0.119 & 0.239 & 0.091 & 0.131
    \end{tabular}
    \caption{Results for \textit{mtool} evaluation of \textbf{R1} and \textbf{R2}.}
    \label{tab:r1_r2_results}
\end{table*}

It must be noted, however, that despite giving us a basis to compare our results to those obtained during the MRP tasks, \textit{mtool} may not be well-suited to evaluate our experiments. To get a better idea of how well our system performs with respect to our goals, we evaluate again with a number of modifications to the UCCA gold data.

To comply with the official UCCA guidelines (see \autoref{subsec:mrp_background}), we evaluate against an updated version of the dataset, where all the punctuation edges (\texttt{U}) have been removed.

AMR annotations do not include anchors. Therefore, without a mapping between the AMR graph and the raw text, we know that producing any would be a guessing game. However, \textit{mtool} takes them into consideration when evaluating UCCA graphs, giving each anchor an equal weight as any edge or node. Thus, anchors constitute a large part of the ``points'' given at evaluation and our system is bound to get lower score because of this. To get a better idea of how well our system does only on nodes and edges, we run an additional evaluation without taking anchors into consideration.

Finally, we put these two modifications together and evaluate the graphs without punctuation and without anchors. 

\autoref{tab:modifications_results} shows the results of these evaluations. As with the \textbf{R2} scores in \autoref{tab:r1_r2_results}, the \textbf{R2} scores here are macro-averages as well. As expected, we get higher scores when punctuation, anchors or both are removed. As seen with the unmodified evaluation, with the exception of precision for \textbf{R2}, we get higher scores on the test set. The \textbf{R2} scores on the train set are significantly higher than those of \textbf{R1} and higher, but by a smaller margin for the test set.

\begin{table*}[]
    \centering
    \begin{tabular}{l||c|c|c||c|c|c}
         &  \multicolumn{3}{c||}{Train} & \multicolumn{3}{c}{Test}\\
         \hline
         & Precision & Recall & F1-score & Precision & Recall & F1-score\\
         \hline
         \hline
         R1 - No punct & 0.128 & 0.040 & 0.061 & 0.179 & 0.062 & 0.092\\
         R1 - No anchors & 0.128 & 0.058 & 0.080 & 0.173 & 0.088 & 0.117\\
         R1 - No punct + no anchors & 0.128 & 0.063 & 0.084 & 0.179 & 0.097 & 0.126\\
         \hline
         R2 - No punct & \textbf{0.280} & 0.100 & 0.147 & \textbf{0.255} & 0.108 & 0.151\\
         R2 - No anchors & 0.249 & 0.126 & 0.167 & 0.239 & 0.147 & 0.181\\
         R2 - No punct + no anchors & \textbf{0.280} & \textbf{0.155} & \textbf{0.198} & \textbf{0.255} & \textbf{0.173} & \textbf{0.204}
    \end{tabular}
    \caption{Results for \textit{mtool} evaluation of the modifications.}
    \label{tab:modifications_results}
\end{table*}

Since with the non-deterministic set of rules, we get a number of output graphs, which differ in at least one edge label from each other, we know that there is one that is closest to the UCCA representation and one that is farthest from it. In \autoref{tab:min_max_avg}, we show again the macro-average of the F1-score of \textbf{R2} and its modifications on the train set and test set, alongside the average of the minimum and the average of the maximum scores for each sentence. In most of the cases, we observe a difference between $0.01$ and $0.02$ on either side of the macro-average.

\begin{table*}[]
    \centering
    \begin{tabular}{l||c|c|c||c|c|c}
         &  \multicolumn{3}{c||}{Train F1-scores} & \multicolumn{3}{c}{Test F1-scores}\\
         \hline
         & Min & Avg & Max & Min & Avg & Max\\
         \hline
         \hline
         R2 & 0.107 & 0.119 & 0.137 & 0.121 & 0.131 & 0.138 \\
         R2 - No punct & 0.132 & 0.147 & 0.167 & 0.143 & 0.151 & 0.159 \\
         R2 - No anchors & 0.150 & 0.167 & 0.191 & 0.168 & 0.181 & 0.192 \\
         R2 - No punct + no anchors & 0.179 & 0.198 & 0.225 & 0.193 & 0.204 & 0.215
    \end{tabular}
    \caption{Minimum, average and maximum F1-scores across train and test set for \textbf{R2} and its modifications.}
    \label{tab:min_max_avg}
\end{table*}

Even though higher than those of \textbf{R1}, the results of \textbf{R2} are still rather low. This is partially due to features of UCCA that cannot be predicted from the AMR only, as we have seen with anchors. However, it is also largely due to ambiguities in the transformation task. We show some examples of these in \autoref{sec:case_study}. These ambiguities stem from the fact that, as one of the six AMR slogans states, we cannot read off a unique English sentence from an AMR\footnote{\url{https://github.com/amrisi/amr-guidelines/blob/master/amr.md\#amr-slogans}}. Thus, producing an UCCA-like representation from AMR is more similar to a generation task. The ambiguities that we describe in \autoref{sec:case_study} can be addressed by adding more non-deterministic rules to the system. This will ensure that we produce a correct graph, but it is not possible to determine which one of the multiple ones produced it is. As the number of output graphs grows exponentially for each non-deterministic rule applied, the task becomes even harder, the more non-deterministic rules we add. This shows that the input graph does not contain enough information to let us compute the correct structure in a deterministic manner.

\section{Ambiguities\label{sec:case_study}}

In this section, we would like to highlight some of the ambiguities that stem from the structural differences of the two frameworks, that we encountered while exploring the train set.

\begin{figure}
  \centering
  \includegraphics[scale=.35]{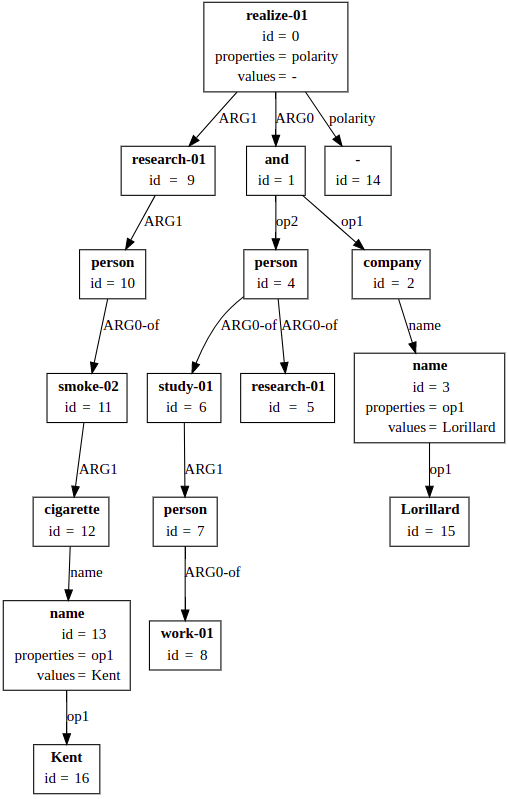}
  \caption{AMR of the sentence ``\emph{Neither Lorillard nor the researchers who studied the workers were aware of any research on smokers of Kent cigarettes.}'' \texttt{[\#20003008]}.}
  \label{lorillard_amr}
\end{figure}

\autoref{lorillard_amr} shows the AMR of sentence \texttt{\#20003008} of the MRP corpus. This is an interesting example for a number of reasons that we have outlined below.

\textbf{Proper names.} In AMR, the structure for annotating a proper names is
\begin{Verbatim}[commandchars=\\\{\},codes={\catcode`$=3\catcode`_=8}]
(e / \textit{entity-type}
    :name (n / name
        :op1 "..."
        ...
        :opN "...")
\end{Verbatim}
 
 where \textit{entity-type} is the type of the entity whose name is used, such as \texttt{person}, \texttt{city}, \texttt{book}\footnote{An exhaustive list of entity types available in AMR can be found in the AMR specifications.} and \texttt{:op1 - :opN} point to each of the tokens in the proper name. In the example in \autoref{lorillard_amr}, we have two such subgraphs - one for \textit{Kent cigarettes} and one for \textit{Lorillard}, which is a \textit{company}. On the surface, however, these are realised in different ways - for \textit{Kent cigarettes} the entity type \texttt{cigarette} is realised along with the name, while for \textit{Lorillard} only the name is present\footnote{Interestingly, this suggests that the AMR graph relies either on context (previous sentences mentioning that Lorillard is a company) or world knowledge. The latter seems to be true for proper names in AMR in general, especially taking into consideration we often include a \texttt{:wiki} relation when a Wikipedia article for that entity is available.}. Thus, in the UCCA representation, the subgraphs for these two instances will have different structures too. It is therefore not possible, from AMR only, without access to the surface realisation of the sentence, to decide whether the entity type should be included in the UCCA representation or not.
 
 \textbf{Nouns that invoke predicates.} Another interesting case is that of AMR's nouns that invoke predicates. In the example from \autoref{lorillard_amr}, we have three such nouns - \textit{researchers}, \textit{workers} and \textit{smokers}. In the AMR graph they are all realised as

\begin{Verbatim}[commandchars=\\\{\},codes={\catcode`$=3\catcode`_=8}]
(p / person
    :ARG\textit{x}-of (p2 / \textit{PB predicate}))
\end{Verbatim}

where \textit{PB predicate} is the relevant PropBank predicate and \textit{x} is the relevant argument number, so e.g. \textit{a smoker} is annotated as \textit{a person who smokes}. This can be addressed by our system by making use of \textsc{Grew}'s lexicons. However, this structure too, is ambiguous. Apart from the three annotations of the three nouns, we have the same structure once more in the example sentence.

\begin{Verbatim}[commandchars=\\\{\},codes={\catcode`$=3\catcode`_=8}]
(p / person
    :ARG0-of (s / study-01))
\end{Verbatim}

Here, however, this does not stand for the noun \textit{student}, but for \textit{[...] who studied}.

\textbf{Negation.} In UCCA, depending on the surface realisation of the sentence, negation can be syntactical (such as \textit{no asbestos} in sentence \texttt{\#20003007}), but also morphological (such as \textit{nonexecutive} in sentence \texttt{\#20001001}). In AMR, negation is marked as \texttt{:polarity -} in both of these cases.

\textbf{have-org-role-91.} Sentences \texttt{\#20001001}, \texttt{\#20001002} and \texttt{\#20003005} all use the special \texttt{have-org-role-91} AMR role  and the same structure when speaking about the organisational roles of specific people. The surface realisations, however, are very different from each other in all three cases - ``\textit{Pierre Vinken [...] will join the board as a nonexecutive director}'', ``\textit{Mr. Vinken is chairman of Elsevier N.V.}'', ``\textit{A Lorillard spokeswoman}''.

\section{Conclusion\label{sec:conclusion}}

In this paper we presented a corpus-driven experiment to transform AMR annotations into UCCA-like representations, the evaluation of our experiment and some of the ambiguous cases we discovered through it. Here we present some of the discussion points stemming from our work and further study directions.

Our work can also be viewed as a case study of seeing how much of an anchored (flavor 1) semantic framework can be predicted from a more abstract (flavor 2) one and what it is that is missing from the latter in order to produce the former. The difficulties in transformation we encountered were largely due to the difference in flavor of the frameworks. UCCA is grounded in surface. As we have seen in \autoref{sec:case_study}, many of the ambiguities would be easier to address if there was a link between AMR and surface as well. This would also help us with predicting where features that are not present in AMR, such as function words, should go in the UCCA-like graph. It would be interesting to see if similar ambiguities arise from comparing other pairs of flavor 1 and 2 frameworks in a similar manner.

In \autoref{sec:evaluation}, we saw that there were a number of adjustments we had to make to the gold dataset in order to get a better idea of how our system performs on the task we set to tackle. Further ones could be made still (such as removing function words). This suggest that \textit{mtool} may not be the most appropriate tool to do such an evaluation. If more experiments in predicting flavor 1 from flavor 2 frameworks (and vice-versa) were to be carried out, there will be the need to design a more appropriate metric to evaluate this kind of task.

Finally, we consider an orthogonal to our task, but equally important issue. Our choice of frameworks was based on the current popularity of the frameworks, but also on the availability of parallel data. Being limited by the second constraint, highlights once again the need for larger and freely available parallel corpora across various semantic frameworks. The availability of a common corpus would greatly enhance corpus-driven comparison across the features and expressive power of various frameworks. Furthermore, whenever a new framework or framework extension is proposed, there would already be a resource that would allow the study of said framework (or extension) with respect to existing ones. Finally, currently the majority of semantically annotated data exists only in English. It would be beneficial if more multi-lingual projects such as the Parallel Meaning Bank \cite{abzianidze-etal-2017-parallel} existed, ideally with datasets that are parallel both across frameworks and languages.

\section{Acknowledgements}

Part of this work has been funded by the \textit{Agence Nationale de la Recherche} (ANR, fr: National Agency for Research), grant number ANR-20-THIA-0010-01. We would like to thank the anonymous reviewers for the time taken to review this paper and provide useful feedback.

\section{Bibliographical References}\label{reference}
\bibliographystyle{lrec2022-bib}
\bibliography{main}
\end{document}